\documentclass[10pt,twocolumn,letterpaper]{article}

\usepackage{cvpr}              
 \usepackage{multirow} 

\usepackage{makecell}
\usepackage{colortbl}
\usepackage{xcolor}

\usepackage{pifont}
\newcommand{\cmark}{\ding{51}}
\newcommand{\xmark}{\ding{55}}

\definecolor{cvprblue}{rgb}{0.21,0.49,0.74}
\usepackage[pagebackref,breaklinks,colorlinks,allcolors=cvprblue]{hyperref}
\usepackage{titletoc}   
\usepackage{fontawesome}
\usepackage{marvosym}

\def\netName{SGDrive}

\title{SGDrive: Scene-to-Goal Hierarchical World Cognition for Autonomous Driving}

\author{
    Jingyu Li$^{1,2*}$ \quad
    Junjie Wu$^{3*}$  \quad
    Dongnan Hu$^{4,2}$ \quad 
    Xiangkai Huang$^{3}$ \quad
    \\
    Bin Sun$^{3\dagger}$ \quad
    Zhihui Hao$^{3\dagger}$ \quad 
    Xianpeng Lang$^{3}$ \quad
    Xiatian Zhu$^{5}$ \quad
    Li Zhang$^{1,2~\textrm{\Letter}}$\quad 
    \\[2mm]
    $^1$ Fudan University \quad
    $^2$ Shanghai Innovation Institute \\
    $^3$ Li Auto Inc. \quad
    $^4$ Tongji University \quad 
    $^5$ University of Surrey \\
    \\ \vspace{-12pt}
    {
    \hypersetup{urlcolor=Cerulean}
    \href{https://github.com/LogosRoboticsGroup/SGDrive}
    {\texttt{{github.com/LogosRoboticsGroup/SGDrive}}}
    }
}

\begin{document}
\maketitle

{\let\thefootnote \relax
\footnote{
\hangindent=1.8em
$^*$Equal contribution.
$^{\dagger}$Project leader. 
$^{~\textrm{\Letter}}$ Corresponding author.\\
This work was done while Jingyu Li and Dongnan Hu were interns at Li Auto Inc.
Primary contact \texttt{jingyuli24@m.fudan.edu.cn}
}}

\begin{abstract}

Recent end-to-end autonomous driving approaches have leveraged Vision-Language Models (VLMs) to enhance planning capabilities in complex driving scenarios. However, VLMs are inherently trained as generalist models, lacking specialized understanding of driving-specific reasoning in 3D space and time. When applied to autonomous driving, these models struggle to establish structured spatial-temporal representations that capture geometric relationships, scene context, and motion patterns critical for safe trajectory planning.
To address these limitations, we propose \netName{}, a novel framework that explicitly structures the VLM's representation learning around driving-specific knowledge hierarchies. Built upon a pre-trained VLM backbone, \netName{} decomposes driving understanding into a scene-agent-goal hierarchy that mirrors human driving cognition: drivers first perceive the overall environment (scene context), then attend to safety-critical agents and their behaviors, and finally formulate short-term goals before executing actions. This hierarchical decomposition provides the structured spatial-temporal representation that generalist VLMs lack, integrating multi-level information into a compact yet comprehensive format for trajectory planning.
Extensive experiments on the NAVSIM benchmark demonstrate that \netName{} achieves state-of-the-art performance among camera-only methods on both PDMS and EPDMS, validating the effectiveness of hierarchical knowledge structuring for adapting generalist VLMs to autonomous driving.
\end{abstract}    
\section{Introduction}
\label{sec:intro}

In recent years, end-to-end~(E2E) autonomous driving techniques have achieved significant strides. However, these methods~\cite{stp3,uniad,vad,sparsedrive} often lack explicit causal reasoning and high-level scene understanding, exhibiting limitations in complex, long-tail traffic scenarios. The advent of large language models~\cite{gpt4,touvron2023llama}, particularly Vision-Language Models~(VLMs)~\cite{internvl,qwen2}, have spurred efforts to integrate their rich prior knowledge and sophisticated reasoning capabilities into the driving tasks, aiming to mitigate these shortcomings and prevent unsafe maneuvers. However, how to effectively translate a VLM's powerful cognitive understanding into physically safe and reliable driving actions remains an open challenge.

\begin{figure}
    \centering
    \includegraphics[width=1.0\linewidth]{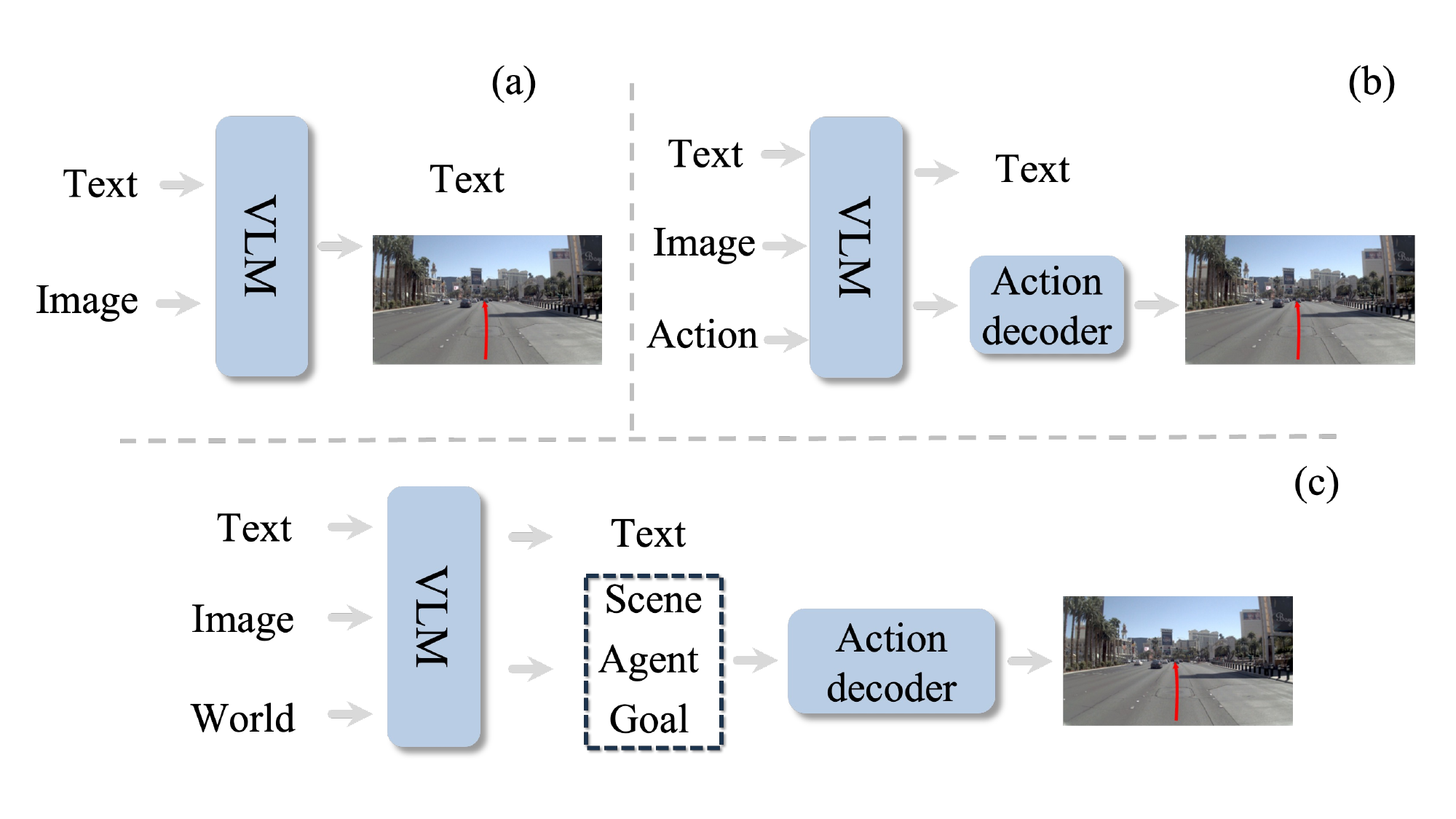}
    \vspace{-10pt}
    \caption{(a)~directly produces driving actions in textual form. (b)~VLM generates action embeddings that decoded to produce the final trajectory. (c)~Our \netName{} explicitly learns and forecasts scene, agent, and goal knowledge, providing structured driving-world understanding that strengthens action reasoning and improves generalization.}
    \label{fig:intro_fig}
    \vspace{-5pt}
\end{figure}
To transfer pretrained VLM knowledge to autonomous driving, several methods~\cite{Drivelm,omnidrive,Drivemm} create domain-specific datasets that adapt the model’s knowledge to driving scenarios.
Building upon this foundation, some works~\cite{EMMA,openemma,ImpromptuVLA} attempt to directly generate trajectories in textual form~Figure\ref{fig:intro_fig}(a). 
Subsequent methods~\cite{ReCogDrive,orion} draw inspiration from embodied intelligence researches~\cite{pi0,openvla} and employ diffusion-based decoder to produce driving trajectories~Figure\ref{fig:intro_fig}(b). 
While the aforementioned methods have achieved impressive results, they still suffer from several key limitations:~(1)~\emph{Lack of spatial perception:} VLMs are inherently focused on semantic understanding and lack foundational spatial and geometric knowledge~\cite{Drivelm,qi2023contrast,depthanything}.~(2)~\emph{Difficulty in discerning critical information}: Current methods~\cite{liu2022petr,bevformer} often focus on the entire scene and lack the extraction of important information, leading to sub-optimal driving performance.~(3)~\emph{Lack of future world state forecast:} Lack of temporal modeling of future world state evolution., such as how the surrounding scene will change. Therefore, we argue that previous methods fail to sufficiently represent the world and forecast its future state, thus hindering the realization of safe and reliable driving.

To address the aforementioned issues, we propose \netName, a novel framework that integrates structured and hierarchical world knowledge into VLMs, thereby enhancing the model's capability to understand and represent driving-relevant world knowledge. 
As shown in Figure\ref{fig:intro_fig}(c), we introduce a set of special tokens, called \textlangle\text{world}\textrangle, which are explicitly trained to extract comprehensive driving-relevant world knowledge. This structured knowledge captures geometric and semantic information as well as high-level driving objectives. By explicitly activating the model's ability to perceive and represent this structured, world knowledge, we fundamentally enhance its 3D spatial perception, enabling the VLM to better guide trajectory generation and avoid potential collisions.

We guide the model to acquire comprehensive driving-relevant world knowledge from three complementary aspects:
(1)~\emph{Scene geometric layout}: Our method leverages occupancy supervision~\cite{occworld,drivingoccwrorld} to learn the overall geometric structure of the scene, enabling the model to perceive and predict holistic spatial variations while removing redundant semantic dependencies.
(2)~\emph{Perceiving driving-relevant agents}: By incorporating an agent detection module~\cite{DETR,liu2022petr}, the model focuses on identifying agents that are likely to influence ego-vehicle motion, instead of all visible objects, aligning better with human driving behavior.
(3)~\emph{Inferring driving goal}: The model learns to reason about feasible driving goal that reflect human-like driving intentions and are consistent with the current scene context.
Together, these components provide the model with a comprehensive prior over both the current and future world states, facilitating safer trajectory planning. 
To ensure disentangled representation across these knowledge, we further apply a block-wise masked attention mechanism to prevent information leakage between different knowledge.
Finally, to effectively translate the driving-relevant world knowledge into trajectory outputs, we employ a diffusion-based transformer~\cite{peebles2023scalable}~(DiT) as the trajectory generator. 
This design enables the model to progressively refine trajectory predictions conditioned on the extracted world knowledge, ensuring coherent trajectory generation.

Our main contributions are summarized as follows: 
(i)~We propose a novel framework that guides the VLM to learn comprehensive world knowledge from different aspects, enhancing its spatial perception and  world representation for safe autonomous driving.
(ii)~We design a block-wise masked attention mechanism to prevent knowledge leakage and noise interference, and couple it with a DiT decoder to generate trajectories conditioned on hierarchical world knowledge.
(iii)~Our method achieves state-of-the-art performance on the NAVSIM benchmark among camera-only methods, demonstrating its effectiveness through extensive experiments.

\section{Related works}
\label{sec:related}

\begin{figure*}[ht]
  \centering
  \vspace{-5pt}
  \includegraphics[width=1.0\textwidth]{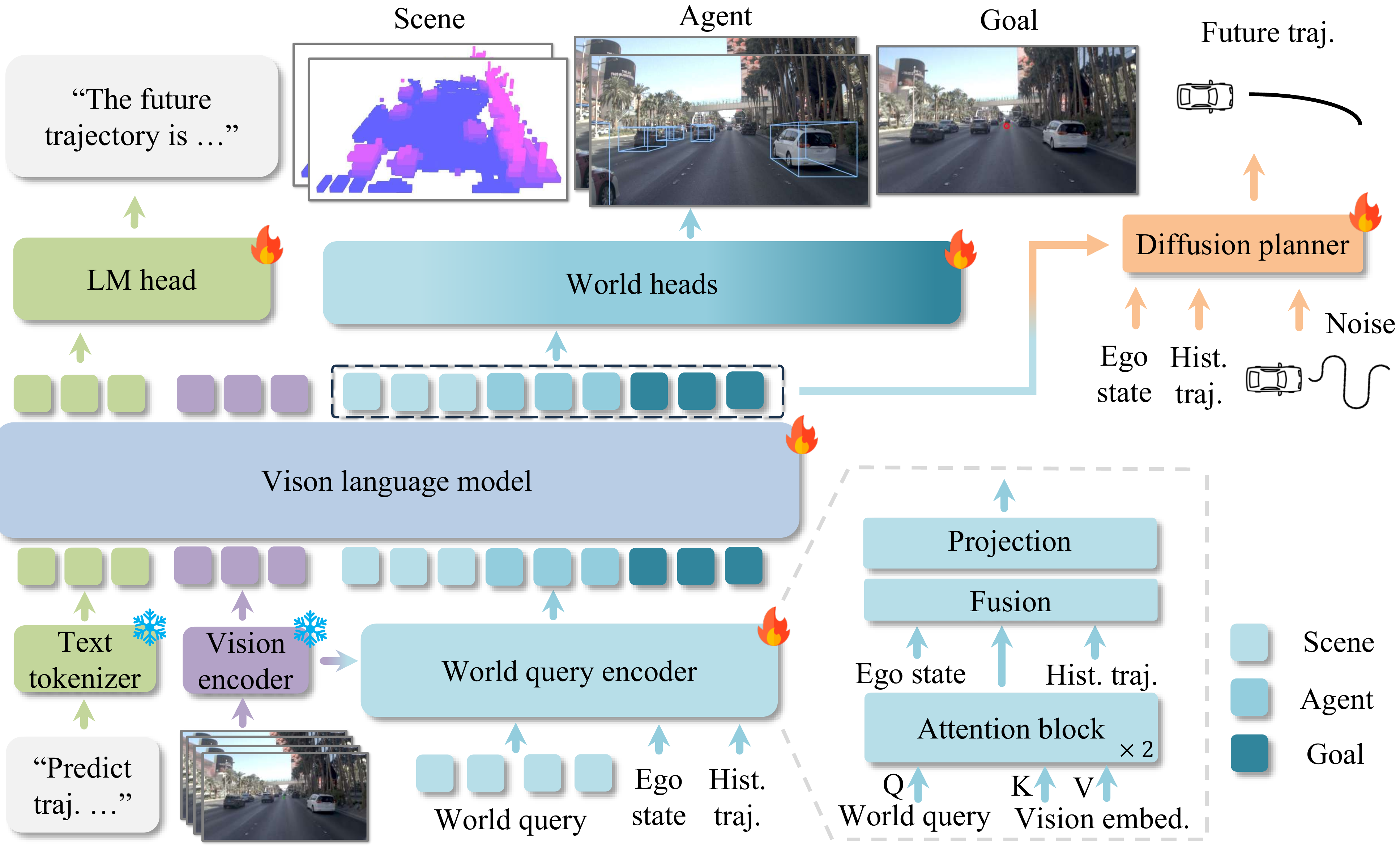} 
  \vspace{-5pt}
  \caption{The \textbf{SGDrive} pipeline introduces hierarchical~\textlangle\text{world}\textrangle\ queries (scene, agent, and goal) for world modeling and trajectory generation. A key component is our world query encoder, which initializes these queries by integrating multi-modal priors from the ego state, historical trajectory, and visual features. These ``prior-informed" queries are then processed by the VLM, alongside text and visual embeddings, to fuse all signals into a compact, hierarchical world representation.}
  \label{fig:pipeline}
  \vspace{-5pt}
\end{figure*}

\subsection{End-to-end autonomous driving}
Recent advancements in autonomous driving have shifted from isolated task pipelines to unified end-to-end planning frameworks that jointly address scene perception and ego-trajectory generation~\cite{stp3,BEVDriver,Mp3,bridgingAD,zhang2025perception}, with further efforts have incorporated multi-modal inputs and transformer-based architectures for enhanced global reasoning~\cite{Transfuser,InterFuser,thinktwice,TF++}. 
UniAD~\cite{uniad} extends the scope by integrating a wide range of subtasks into a cohesive system, while VAD~\cite{vad} further improves it with vectorized representations and refined modular design. 
SparseAD~\cite{sparsead} and SparseDrive~\cite{sparsedrive} explore sparse representations to improve the efficiency and scalability of end-to-end planning systems.
VADv2~\cite{vadv2} pioneers the integration of probabilistic planning into end-to-end autonomous driving.
DiffusionDrive~\cite{diffusiondrive} leverages a truncated diffusion policy to improve the accuracy and diversity of planned trajectories. 
WoTE~\cite{WoTE} builds a world-model-based autonomous driving system upon the BEV representation.
Although these methods have achieved remarkable success, their imitation-learning nature inherently limits generalization to long-tail and complex scenarios, while lacking reasoning and deliberative capabilities.

\subsection{Vision-language-model in autonomous driving}

Within autonomous driving, LLMs~\cite{gpt4} and VLMs~\cite{BLIP-2, Minigpt-4, LLaVA} have revealed strong reasoning and multi-modal capabilities. 
Concurrently, the introduction of large-scale driving datasets~\cite{Bench2Drive,carla,nuPlan} with natural language annotations~\cite{Reason2drive,Nuscenes-qa,marcu2024lingoqa,zhou2025hermes} has facilitated research.
Notably, DriveLM~\cite{Drivelm} introduces a Chain-of-Thought reasoning paradigm spanning perception to planning, while DriveMM~\cite{Drivemm} integrates diverse language-driving datasets to build a universal VLM. 
Inspired by advancements in the field of robotic manipulation~\cite{openvla,pi0,DreamVLA}, a new class of Vision-Language-Action (VLA) models based on VLMs has emerged to autonomous driving~\cite{drivevlm,senna,li2025imagidrive,liu2025drivepi}.
EMMA~\cite{EMMA}, built upon Gemini~\cite{gemini}, maps raw camera inputs to driving decisions and demonstrates high accuracy in perception and strong performance in motion planning. 
SimLingo~\cite{simlingo} introduces a VLM-based architecture that unifies driving, vision-language understanding, and language-action alignment.  
OpenDriveVLA~\cite{opendrivevla} employs hierarchical vision-language alignment and agent-environment-ego interaction to produce reliable driving actions, excelling in planning and driving-related question answering tasks.
ORION~\cite{orion} proposes a framework that integrates QT-Former, a large language model (LLM), and a generative planner, achieving strong performance in closed-loop evaluations.
FSDrive~\cite{FutureSightDrive} reformulates CoT reasoning into a spatio-temporal visual CoT, where a VLM generates a future frame and subsequently predicts the trajectory via inverse dynamics.
ReCogDrive~\cite{ReCogDrive} proposes a framework that integrates a VLM with a diffusion planner, utilizing Diffusion Group Relative Policy Optimization to enhance trajectory generation.
In contrast to previous work, \netName{} represents and forecasts world knowledge in a hierarchical (scene-agent-goal) and effective manner, enabling safe and reliable driving. 

\section{Method}

\subsection{Problem definition and notation}
We aim to enhance the safety of autonomous driving through the guidance of multi-level driving-relevant world knowledge. Accordingly, we formulate our framework as tackling two complementary sub-problems: extracting representative world knowledge and extrapolating future world states. At each time step~$t$, the ego-vehicle receives heterogeneous signals: a natural language instruction~$L_{ins}$, the ego-vehicle state~$S_{ego}$, and camera sensor inputs~$I_{cam}$. These inputs together constitute the world knowledge of the driving system. To direct the model’s attention toward driving cues and facilitate forecasting of future world states, we introduce a set of special tokens, denoted as~\textlangle\text{world}\textrangle\ , and adopt a VLM to transform the comprehensive world knowledge into compact latent representations:
\begin{equation}
    O_{\text{world}} = 
    VLM\bigl(I_{cam},\, L_{ins},\, S_{ego}\,|\,\textlangle\text{world}\textrangle\bigr).
\end{equation}
The resulting representation $O_{\text{world}}$ encapsulates the driving-related world knowledge and serves as the foundation for forecasting the future evolution of the world state.
We then introduce a set of hierarchical world heads~$\mathcal{D}$ that extract structured world knowledge across geometric details, motion cues, and high-level cognition, and further extrapolate the world state at future time~$t{+}n$:
\begin{equation}
\label{eq:world_decoding}
w = \mathcal{D}\bigl(O_{world}\bigr) = \{w^{t,t+n}_{geo},\, w^{t,t+n}_{agt},\, w_{goal}\},
\end{equation} 
where $w_{geo}$ represents the layout of the scene, $w_{agt}$ captures the state of the safety-critical agents, and $w_{goal}$ encodes the short-term driving objective. 

Based on the learned world knowledge, we utilize DiT~\cite{peebles2023scalable} to generate trajectory. This design effectively translates the forecast world knowledge into coherent and safe future trajectories, completing the reasoning chain from scene understanding to motion planning.

\subsection{Model architecture}

As illustrated in Figure~\ref{fig:pipeline}, our \netName~ builds on a foundational VLM with two core modules. \netName ~accepts heterogeneous inputs and processes each modality separately. Specifically, we use a standard text tokenizer encodes the driving instruction, while a ViT-based visual encoder~\cite{InternVL3} extracts features from the camera image. We further introduce a set of \textlangle\text{world}\textrangle\  queries, which are appended to the multimodal embeddings. And these queries consist of three subqueries~(scene, agent, and goal), which is used for predicting hierarchical driving world knowledge.
The world queries are initialized via a world query encoder, which integrates multi-modal priors from the ego state, historical trajectory, and visual embeddings (Figure~\ref{fig:pipeline}). These prior-informed queries effectively capture contextual information from the scene. Combined with the VLM’s strong priors, our method fuses visual and ego-vehicle signals into a compact hierarchical representation that encodes the scene geometry, dynamic agent states, short-term driving objectives, and predicted future world states, providing a structured and interpretable foundation for subsequent world modeling and trajectory generation.

In the decoding stage, task-specific heads process the corresponding subqueries, transforming the unified world embedding into explicit representations~(scene context, agent states, and short-term goals). The latent world embedding implicitly conveys hierarchical driving knowledge, enabling downstream trajectory generation without requiring explicit decoding of the \textlangle\text{world}\textrangle\ queries, thereby reducing computational cost while preserving semantic richness.

\subsection{Hierarchical world knowledge representation}
\label{sec:hwkr}
Safe driving requires anticipating how the surrounding environment will evolve. Our method mirrors human driving cognition by explicitly forecasting future world knowledge along three complementary aspects: scene geometry, safety-critical agents, and short-term driving goals. This structured scene-agent-goal representation supplements the missing depth information in images and provides predictive context to support safe trajectory generation.

\noindent\textbf{Geometric scene layout perception.}
Perceiving and forecasting the geometric layout of the driving scene provides essential spatial cues for safe driving. Our model does not predict high-level semantic distributions; instead, it focuses on the overall geometric structure. When occupancy annotations are available in the dataset, we supervise the model with ground-truth labels; otherwise, we generate occupancy from the point clouds.
Since a VAE decoder excels at reconstructing features from latent representations~\cite{VQVAE,Occ3d,OccVLA}, we treat the VLM output~$W_{\text{geo}}$ as the latent embedding and employ a standard VAE decoder for geometric reconstruction. Considering the high sparsity of driving scenes with a large number of negative samples~\cite{OccVLA}, we follow prior work~\cite{song2017semantic} and adopt a resampling strategy, supervised by two classification losses to ensure balanced learning of occupied and unoccupied regions:
\begin{align}
\mathcal{L}_{\text{geo}}^{t,t+n} &= 
\frac{1}{M} \sum_{i=1}^{M} 
\mathrm{CE}(o_i^{t,t+n}, \hat{o}_i^{t,t+n})\notag\\
&\quad+ 
\frac{1}{N} \sum_{j=1}^{N} 
\mathrm{BCE}(p_j^{t,t+n}, \hat{p}_j^{t,t+n}),
\end{align}
where $\mathcal{L}_{\text{geo}}$ combines the standard cross-entropy loss over all spatial locations with a resampled binary cross-entropy term to handle sparse occupancy distribution. $o_i  \in \{0,1\}$ indicates whether location $j$ is occupied in the ground truth, and $p_i$ denotes resampled candidate positions.

\noindent\textbf{Safety-critical agents detection.} 
To handle complex driving interactions, our \netName~focuses on safety-critical road users that directly influence driving behavior, encouraging collision-avoidant reasoning and deeper understanding of safety-critical interactions. 
Specifically, we select target agents~(vehicle, pedestrian and cyclist) based on ego-vehicle trajectory and visibility from the front-view camera frustum. 
This strategy compels the model to allocate its finite representational capacity to agents most relevant to the ego-vehicle’s decisions, rather than exhaustively perceiving all objects in the scene. For these selected agents, the model predicts their 3D states at both the current and future time steps~($t$ and $t+n$).
To supervise these agent predictions, we adopt the set-based loss paradigm from DETR~\cite{DETR}, which finds an optimal bipartite matching $\hat{\sigma}$ between the $N_q$ predictions and the set of ground-truth objects. The total loss $\mathcal{L}_{\text{agent}}$ is then computed as a weighted sum of classification and regression losses for the matched pairs:
\begin{align}
\mathcal{L}_{\text{agent}}^{t,t+n} &= \sum_{i=1}^{N_q} \bigl[ \lambda_{\text{cls}} \mathcal{L}_{\text{cls}}(\hat{c}_i^{t,t+n}, c_{\hat{\sigma}(i)}^{t,t+n}) \notag\\
&\quad+ \mathbf{1}_{c_{\hat{\sigma}(i)^{t,t+n}} \neq \emptyset} \mathcal{L}_{\text{reg}}(\hat{b}_i^{t,t+n}, b_{\hat{\sigma}(i)}^{t,t+n}) \bigr],
\end{align}
where $\hat{\sigma}(i)$ is the index of the ground-truth object matched to the $i$-th prediction. $\mathcal{L}_{\text{cls}}$ is a cross-entropy loss, $\lambda_{\text{cls}}$ takes 10, and $\mathcal{L}_{\text{reg}}$ is an $L_1$ loss. The indicator term $\mathbf{1}_{c_{\hat{\sigma}(i)} \neq \emptyset}$ ensures the regression loss is applied only to positive matches.

\noindent\textbf{Short-term driving goal forecasting.}
Predicting short-term driving goals provides high-level semantic guidance for the ego-vehicle, indicating the intended trajectory over the immediate future. Without such predictions, ego-vehicle may exhibit incomplete or suboptimal maneuvers, such as covering only part of the planned path—potentially reducing task efficiency and safety.
At the apex of the cognitive hierarchy, multi-modal interactions between visual and textual embeddings are used to infer the ego-vehicle's intended objective. Rather than being directly conditioned on the previously established world representations (e.g., scene geometric layout or safety-critical agents), this goal reasoning emerges implicitly from a holistic understanding of the scene and task instructions.
Building upon this implicit goal reasoning, \netName{} predicts a short-term driving goal, $\hat{p}_{\text{goal}}$, defined as the target ego-pose approximately 4 seconds into the future. This prediction is decoded via a lightweight mlp head and supervised using an $L_1$ loss against the ground-truth pose $p_{\text{goal}}$:
\begin{equation}
\mathcal{L}_{\text{goal}} = || \hat{p}_{\text{goal}} - p_{\text{goal}} ||_1.
\end{equation}
The predicted goal is more than a single trajectory point; it encodes high-level driving intentions informed by the model’s understanding of scene constraints and potential interactions. By explicitly predicting this goal, we effectively disentangle high-level decision-making from low-level trajectory planning.

\begin{figure}
  \centering
  \includegraphics[width=1.0\linewidth]{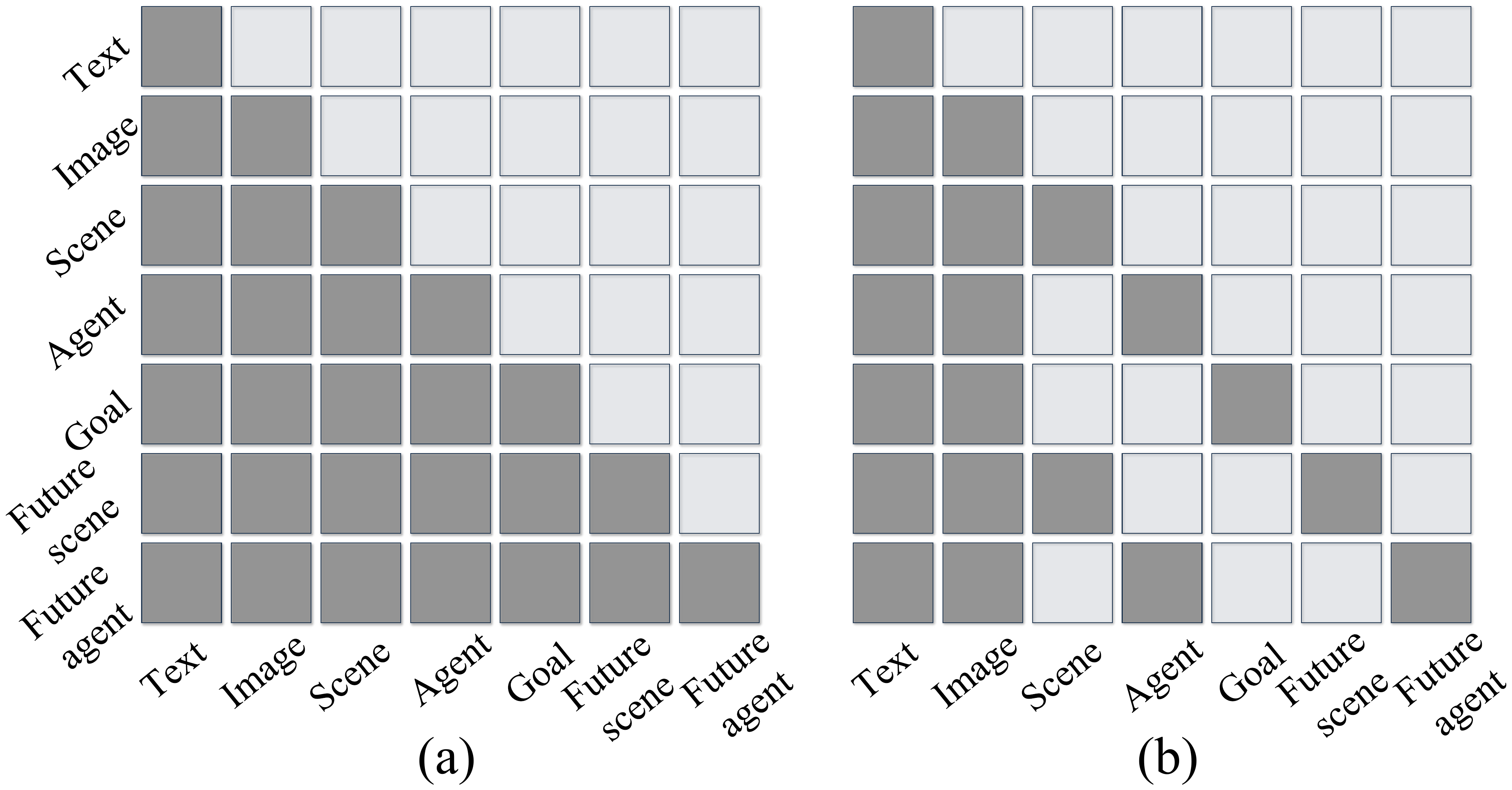}
  \vspace{-10pt}
  \caption{
(a)~Causal attention mask: input tokens are allowed to attend to tokens before.
(b)~Structure attention mask: prevents leakage by prohibiting all mutual attention between the different subquery sets~(scene, agent, goal).
  }
  \label{fig:attention}
\end{figure}
\noindent\textbf{Structured block-wise attention mask for driving-world knowledge.}  
Our \netName{} employs specialized \textlangle\text{world}\textrangle\ queries to decode anticipatory knowledge, capturing both current state and future evolution. A key challenge in this multi-task design is representational contamination: if queries freely attend to each other (Figure~\ref{fig:attention}(a)), information can leak across cognitive levels, compromising the integrity of specialized representations.  

To address this, we introduce a block-wise structured attention mask (Figure~\ref{fig:attention}(b)). The \textlangle\text{world}\textrangle\ queries are divided into five subqueries: the first three encode current world knowledge, and the remaining two focus on forecasting future states. The mask blocks attention across different knowledge categories while allowing temporal attention within each category, enabling subqueries to access relevant historical context. All subqueries remain free to cross-attend to the primary input modalities (visual and text embeddings), ensuring each representation gathers necessary evidence.  
This structured attention effectively prevents cross-level leakage, maintaining specialized and accurate hierarchical representations, which is essential for safe and precise driving.

\subsection{Diffusion planner}
A key challenge in autonomous driving is bridging the gap between high-level semantic reasoning and low-level continuous actions. To address this, we employ a diffusion planner~\cite{openvla,pi0,peebles2023scalable} that generates safe and context-aware action trajectories. Our \textlangle\text{world}\textrangle\ queries, which encode a hierarchical and anticipatory understanding of the driving world, are directly used as the latent condition for the planner. This avoids intermediate, lossy representations and enables access to the VLM's full world knowledge while reducing inference overhead.

The planner denoises a sequence of future waypoints $\mathbf{A} = (a_1, \dots, a_N)$ from a noisy initialization $\mathbf{A}_T$ to the ground-truth trajectory $\mathbf{A}_0$ over $T$ steps. The per-step denoising is conditioned on the hierarchical world knowledge and ego-vehicle state, injected via the DiT's cross-attention layers. Instead of starting from pure Gaussian noise, $\mathbf{A}_T$ is initialized by adding noise $\epsilon$ to a learned prior, generated from a linear projection of the \textlangle\text{world}\textrangle\ queries and historical ego-trajectory, grounding the process in the VLM's world understanding. The diffusion model $\epsilon_\theta$ is trained with a standard $L_2$ objective:
\begin{equation}
\mathcal{L}_{\text{diff}} = \mathbb{E}_{t, \mathbf{A}_0, \epsilon} \left[ \left\| \epsilon - \epsilon_\theta(\mathbf{A}_t, t, c) \right\|_2^2 \right],
\end{equation}
where $\mathbf{A}_t$ is the noisy trajectory at step $t$ and $c$ denotes the per-step condition.

\begin{table*}[ht]
    \centering
    \vspace{-5pt}
    \small
    \caption{Performance comparison on NAVSIM v1 \textit{navtest} using closed-loop metrics. We additionally report results with reinforcement learning fine-tuning~(RFT) to enable fair comparison with methods that adopt such training strategies. $^{\dagger}$ denotes models fine-tuned on the NAVSIM trajectory dataset. Bold indicates the best results under SFT and RFT settings, respectively.}
    \vspace{-5pt}
    \setlength{\tabcolsep}{7pt}
    \begin{tabular}{@{}l|cc|cc|ccc|cc@{}}
        \toprule
        Method &  Image & Lidar & NC$\uparrow$ & DAC$\uparrow$ & TTC$\uparrow$ & Comf. $\uparrow$ & EP$\uparrow$ & PDMS$\uparrow$ \\
        \midrule
        Constant Velocity & &   & 68.0 & 57.8 & 50.0 & 100 & 19.4 & \cellcolor{gray!30} 20.6 \\
        Ego Status MLP &  &  & 93.0 & 77.3 & 83.6 & 100 & 62.8 & \cellcolor{gray!30} 65.6 \\        
        \midrule
        VADv2-$\mathcal{V}_{\text{8192}}$~\citep{vadv2} & \checkmark &  &  97.2 & 89.1 & 91.6 & 100 & 76.0 &\cellcolor{gray!30}  80.9 \\ 
        Hydra-MDP-$\mathcal{V}_{\text{8192}}$~\citep{Hydra-mdp} & \checkmark & \checkmark &   97.9 & 91.7 & 92.9 & 100 & 77.6 & \cellcolor{gray!30} 83.0 \\
        UniAD~\citep{uniad} & \checkmark &    & 97.8 & 91.9 & 92.9 & 100 & 78.8 & \cellcolor{gray!30} 83.4 \\
        LTF~\citep{Transfuser} & \checkmark &    & 97.4 & 92.8 & 92.4 & 100 & 79.0 & \cellcolor{gray!30} 83.8 \\
        BevDrive~\citep{BEVDriver} & \checkmark & \checkmark   & 97.7 & 92.5 & 92.9 & 100 & 78.7 & \cellcolor{gray!30} 83.8 \\
        TransFuser~\citep{Transfuser} & \checkmark & \checkmark   & 97.7 & 92.8 & 92.8 & 100 & 79.2 & \cellcolor{gray!30} 84.0 \\
        PARA-Drive~\citep{para-drive} & \checkmark &  & 97.9 & 92.4 & 93.0 & 99.8 & 79.3 & \cellcolor{gray!30} 84.0 \\
        DRAMA ~\citep{drama} & \checkmark & \checkmark   & 98.0 & 93.1 & 94.8 & 100 & 80.1 & \cellcolor{gray!30} 85.5 \\
        Epona~\citep{Epona} & \checkmark &    & 97.9 & 95.1 & 93.8 & 99.9 & 80.4 & \cellcolor{gray!30} 86.2 \\
        Hydra-MDP-$\mathcal{V}_{\text{8192}}$-W-EP~\citep{Hydra-mdp} & \checkmark & \checkmark  & 98.3 & 96.0 & 94.6 & 100 & 78.7 & \cellcolor{gray!30} 86.5 \\
        ARTEMIS~\citep{ARTEMIS} & \checkmark & \checkmark & 98.3 & 95.1 & 94.3 & 100 & 81.4 & \cellcolor{gray!30} 87.0 \\
        DiffusionDrive~\citep{diffusiondrive} & \checkmark & \checkmark & 98.2 & 96.2 & 94.7 & 100 & 82.2 & \cellcolor{gray!30} 88.1 \\
        WoTE~\citep{WoTE} & \checkmark & \checkmark & 98.5 & 96.8 & 94.9 & 99.9 & 81.9 & \cellcolor{gray!30} 88.3 \\
        SeerDrive~\citep{zhang2025future} & \checkmark & \checkmark & 98.4 & 97.0 & 94.9 & 99.9 & 83.2 & \cellcolor{gray!30} 88.9 \\
        \midrule
        \multicolumn{8}{@{}l}{\raggedright \textbf{VLMs-based Methods~(SFT)}} \\
        AutoVLA-3B~\citep{AutoVLA} & \checkmark &  &  96.9 & 92.4 & 88.1 & 99.1 & 75.8 & \cellcolor{gray!30} 80.5 \\
        QwenVL2.5-8B$^{\dagger}$~\citep{qwen25vl} & \checkmark &  &  97.8 & 92.1 & 92.8 & \textbf{100} & 78.3 & \cellcolor{gray!30} 83.3 \\
        InternVL3-8B$^{\dagger}$~\citep{InternVL3} & \checkmark &  &  97.0 & 92.4 & 91.8 & \textbf{100} & 78.9 & \cellcolor{gray!30} 83.3 \\
        
        ReCogDrive-2B~\citep{ReCogDrive} & \checkmark &  &  98.1 & 94.7 & 94.2 & \textbf{100} & 80.9 & \cellcolor{gray!30} 86.5 \\
        ReCogDrive-8B~\citep{ReCogDrive} & \checkmark &  &  98.3 & 95.1 & 94.3 & \textbf{100} & 81.1 & \cellcolor{gray!30} 86.8 \\
        \midrule
        \rowcolor{gray!30}SGDrive-2B~(ours) & \checkmark &  & \textbf{98.6} & \textbf{95.1} & \textbf{95.4} & \textbf{100} & \textbf{81.2} & \cellcolor{gray!30} \textbf{87.4} \\
        \midrule
        \multicolumn{8}{@{}l}{\raggedright \textbf{VLMs-based Methods~(RFT)}} \\
        AutoVLA-3B~\citep{AutoVLA} & \checkmark &  &  98.4 & 95.6 & \textbf{98.0} & 99.9 & 81.9 & \cellcolor{gray!30} 89.1 \\
        ReCogDrive-2B~\citep{ReCogDrive} & \checkmark &  &  97.9 & 97.3 & 94.9 & \textbf{100} & \textbf{87.3} & \cellcolor{gray!30} 90.8 \\
        ReCogDrive-8B~\citep{ReCogDrive} & \checkmark &  &  97.8 & 97.7 & 94.9 & \textbf{100} & 86.3 & \cellcolor{gray!30} 90.5 \\
        \midrule
        \rowcolor{gray!30}SGDrive-2B~(ours) & \checkmark &  & \textbf{98.6} & \textbf{97.8} & 96.2 & \textbf{100} & 85.8 & \cellcolor{gray!30} \textbf{91.1} \\
        \bottomrule
    \end{tabular}
    \label{tab:main_results_on_pdms_rl}
    \vspace{-5pt}
\end{table*}

\subsection{Training objectives}
Our \netName{} is trained in a two-stage procedure to effectively manage the distinct task spaces of world representation and action generation. 
In the first stage, we perform Supervised Fine-Tuning~(SFT) to train the core VLM for both visual question answering~(VQA) and comprehensive world knowledge acquisition. The model processes multi-frame front-view camera inputs, ego-vehicle states, and language-based driving commands, and is supervised to jointly predict: (1)~textual answers for VQA~$\mathcal{L}_{text}$, (2)~the spatio-temporal scene geometry layout, (3)~safety-critical agent detection, and (4)~the short-term driving goal. The total loss is defined as:
\begin{equation}
\mathcal{L}_{\text{Stage1}} = 
\mathcal{L}_{\text{text}} + 
\mathcal{L}_{\text{occ}}^{t,t+n} + 
\lambda_{\text{agent}}\mathcal{L}_{\text{agent}}^{t,t+n} + 
\mathcal{L}_{\text{goal}},
\end{equation}
where $\lambda_{\text{agent}}$ takes $0.1$.

In the second stage, we freeze the pre-trained VLM from stage~1 to serve as a high-fidelity world model, and train the diffusion planner with the same inputs but a single optimization target, the trajectory diffusion loss $\mathcal{L}_{\text{diff}}$. This staged strategy enables the VLM to first learn a robust, general-purpose representation of the driving world, which is then exploited by the diffusion planner to generate safe and realistic trajectories.

\section{Experiments}
\begin{table*}[ht]
  \centering
  \caption{Performance comparison on NAVSIM v2 \textit{navtest} with extended metrics. Our SGDrive-2B is evaluated using the model trained with the proposed two-stage SFT strategy.}
  \label{tab:main_results_on_epdms}
  \small
  \scalebox{1}{
  \begin{tabular}{l|ccccccccc|c}
    \toprule
    Method     & NC$\uparrow$ & DAC$\uparrow$ & EP$\uparrow$ & TTC$\uparrow$ & HC$\uparrow$ & TL$\uparrow$ & DDC$\uparrow$ & LK$\uparrow$ & EC$\uparrow$ & EPDMS$\uparrow$ \\
    \midrule
    Transfuser~\citep{Transfuser}    & 97.7 & 92.8 & 79.2 & 92.8 & \textbf{100}   & 99.9 & 98.3 & 67.6 & 95.3 & \cellcolor{gray!30}77.8 \\
    VADv2~\citep{vadv2}         & 97.3 & 91.7 & 77.6 & 92.7 & \textbf{100}   & 99.9 & 98.2 & 66.0 & 97.4 & \cellcolor{gray!30}76.6 \\
    Hydra-MDP~\citep{Hydra-mdp}     & 97.5 & 96.3 & 80.1 & 93.0 & \textbf{100}   & 99.9 & 98.3 & 65.5 & 97.4 & \cellcolor{gray!30}79.8 \\
    Hydra-MDP++~\citep{Hydra-mdp}   & 97.9 & \textbf{96.5} & 79.2 & 93.4 & \textbf{100}   & \textbf{100.0} & 98.9 & 67.2 & \textbf{97.7} & \cellcolor{gray!30}80.6 \\
    ARTEMIS~\citep{ARTEMIS}           & 98.3 & 95.1 & 81.5 & 97.4 & \textbf{100} & 99.8 & 98.6 & 96.5 & 98.3 & \cellcolor{gray!30}83.1 \\
    ReCogDrive-8B~\citep{ReCogDrive}  & 98.3 & 95.2 & 87.1 & 97.5 & 98.3 & 99.8 & \textbf{99.5} & 96.6 & 86.5 & \cellcolor{gray!30}83.6 \\
    DiffusionDrive~\citep{diffusiondrive}  & 98.0 & 96.0 & \textbf{87.7} & 97.1 & 98.3 & 99.8 & \textbf{99.5} & \textbf{97.2} & 87.6 & \cellcolor{gray!30}84.3 \\
    \midrule
    SGDrive-2B~(ours)  & \textbf{98.6} & 94.3 & 86.0 & \textbf{97.9} & 98.3 & 99.9 & \textbf{99.5} & 96.1 & 85.9 & \cellcolor{gray!30}\textbf{86.2} \\
    \bottomrule
  \end{tabular}}
\end{table*}
\subsection{Experimental setup}  
\noindent\textbf{Implementation Details.}
We use InternVL3-2B~\citep{InternVL3} as our VLM backbone, which is integrated a 300M-parameter InternViT visual encoder~\citep{internvl} with the Qwen2.5 large-language-model~\citep{qwen25vl}.
In stage 1, we first perform domain adaptation to align the base VLM with the driving modality, following~\cite{ReCogDrive}. We use over 3.1 million question-answer~(QA) pairs covering perception, prediction, and planning, and train for 1 epoch. Subsequently, we fine-tune the VLM on 85k trajectory-specific QA pairs while concurrently training the world knowledge heads for 3 epoches.  
In Stage 2, we freeze the VLM parameters and train the diffusion planner exclusively for 220 epochs. All experiments are conducted on 4 nodes, each equipped with 8 NVIDIA H20 GPUs (32 GPUs total). Additional details are provided in the supplementary material.

\noindent\textbf{Dataset and evaluation metrics.}
NAVSIM~\cite{navsim} is a large-scale real-world autonomous driving dataset designed for non-reactive simulation and benchmarking. It focuses on challenging scenarios involving dynamic intention changes, while filtering out trivial cases such as stationary or constant-speed driving. It is split into two subsets: navtrain (1,192 scenarios) for training and validation, and navtest (136 scenarios) for testing.
As for the evaluation metrics, we evaluate our method using the Predictive Driver Model Score~(PDMS) and the Extended PDMS~(EPDMS), as defined in the official benchmark. PDMS consists of several sub-scores, including No At-Fault Collisions~(NC), Drivable Area Compliance~(DAC), Time-to-Collision~(TTC), Comfort~(Comf.), and Ego Progress~(EP). EPDMS further introduces Traffic Light Compliance~(TL), Lane Keeping Ability~(LK), and Extended Comfort~(EC), providing a more comprehensive evaluation.

\begin{figure*}
  \centering
  \vspace{-5pt}
  \includegraphics[width=1.0\textwidth]{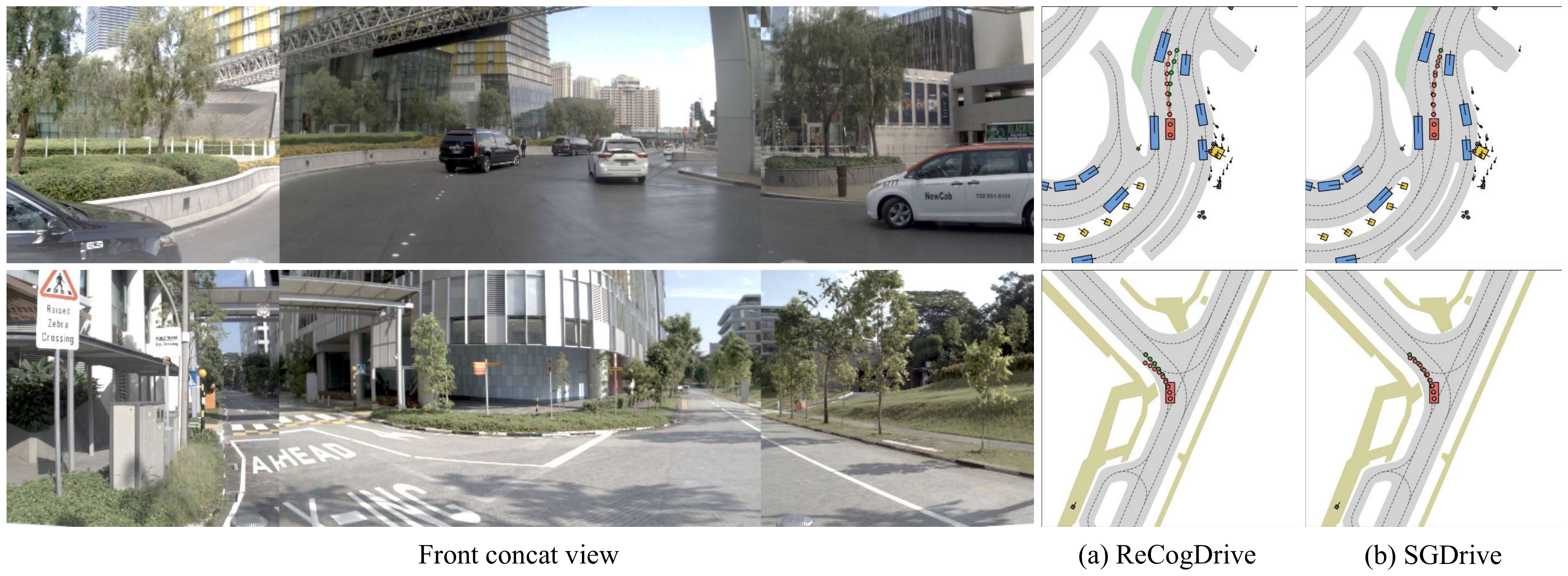} 
  \vspace{-15pt}
  \caption{Comparisons with state-of-the-art method on the Navtest benchmark.}
  \label{fig:comparison_on_diff_model}
\end{figure*}

\subsection{Main result.}
\noindent\textbf{Results on NAVSIM v1 with SFT.} 

As shown in Table~\ref{tab:main_results_on_pdms_rl}, we compare \netName{} against state-of-the-art approaches on the NAVSIM test split. Our approach, built upon an InternVL3-2B backbone and trained using our proposed two-stage supervised fine-tuning strategy, achieves a new state-of-the-art PDMS score of 87.4.
This result is notable for several reasons. First, it surpasses larger general-purpose VLMs like InternVL3-8B and QwenVL2.5-8B by a significant 4.1 PDMS, demonstrating the superior performance of our specialized architecture. Second, \netName{}-2B model outperforms the previous state-of-the-art driving VLM method, Recogdrive-8B~\cite{ReCogDrive}, by 0.6 PDMS. This highlights the profound effectiveness of guiding the VLM to learn and forecast hierarchical world knowledge, as this enables a more compact model to achieve superior planning performance. Third, \netName{} relay on image-only input outperforms the vast majority of listed end-to-end methods that rely on both image and LiDAR inputs.

Crucially, our method achieves the best scores on the key collision-related metrics, NC and TTC. This strongly validates our core hypothesis: by explicitly forecasting the spatio-temporal layout, dynamic agent interactions, and short-term goals, the model gains a superior spatial-temporal awareness that is paramount for anticipating and avoiding potential collisions.

\noindent\textbf{Results on NAVSIM v1 with RFT.}
Although our method primarily aims to learn hierarchical driving-world knowledge to improve driving safety, it can be seamlessly integrated with existing RL frameworks. Under the same RL training configuration as RecogDrive~\cite{ReCogDrive}, our approach achieves substantially better results, as shown in Table~\ref{tab:main_results_on_pdms_rl}. By incorporating structured world knowledge features into the RL pipeline, our method achieves a PDMS of 91.1, outperforming all existing methods, including those using LiDAR inputs.
Compared with other RL-based approaches, our model achieves the best performance on NC and DAC, indicating that the learned driving-world knowledge effectively reduces collision risk and improves compliance with drivable regions. In future work, we plan to explore RL algorithms specifically tailored to our hierarchical world knowledge forecasting framework to further improve driving efficiency and smoothness.

\noindent\textbf{Results on NAVSIM v2 with SFT.}
To comprehensively evaluate our approach, we also follow prior work~\cite{Hydra-mdp} and adopt the Extended PDMS metric on the NAVSIM~\cite{navsim} benchmark. As shown in Table~\ref{tab:main_results_on_epdms}, \netName{} achieves the best overall performance with an EPDMS of~86.2, outperforming the previous state-of-the-art ReCogDrive-8B by~2.6 points. Our method also delivers the strongest results on the safety-critical NC and TTC metrics, while maintaining competitive performance on the newly introduced TL, LK, and EC metrics. These results collectively demonstrate the effectiveness and robustness of \netName{} in modeling driving-relevant world knowledge under the extended evaluation protocol.

\begin{table}[t]
  \centering
  \small
  \caption{Ablation study on the proposed components of SGDrive.}
  \setlength{\tabcolsep}{2pt}
  \vspace{-5pt}
  \begin{tabular}{
    c
    c c c c c c c
    c c c c c | c c
  }
    \toprule
    Exp. 
      & Base
      & Current
      & Future
      & NC$\uparrow$  
      & DAC$\uparrow$ 
      & TTC$\uparrow$ 
      & EP$\uparrow$  
      & PDMS$\uparrow$  \\
    \midrule
    a 
      & \cmark & \xmark & \xmark  & 97.3 & 91.1 & 92.9 & 76.8 & \cellcolor{gray!30} 82.2  \\
    b 
      & \cmark & \cmark & \xmark
      & 98.3 & 93.0 & 94.9 & 78.2 & \cellcolor{gray!30} 84.7  \\
    c 
      & \cmark & \cmark & \cmark 
       & 98.4 & 93.6 & 94.9 & 79.3& \cellcolor{gray!30} 85.5  \\
    \bottomrule
  \end{tabular}
  \vspace{-5pt}
  \label{tab:ablation_results}
\end{table}

\subsection{Ablation study}
\noindent\textbf{Effect of our driving-world knowledge forecast.}

We first evaluate the effectiveness of our proposed driving-world knowledge learning in stage~1, where trajectories are produced in text form and the result is shown in Table~\ref{tab:ablation_results}. When the model is trained only to represent the multi-level structure of the \emph{current} world state, as in Exp.(b), it achieves a 2.5 points improvement in PDMS over Exp.(a). This notable gain demonstrates that our hierarchical world representation successfully activates the model's understanding of the 3D driving environment, leading to more accurate trajectory predictions.
When we further incorporate \emph{future} world forecasting in Exp.(c), the performance increases to 85.5~PDMS, along with additional improvements in the NC and EP metrics compared with Exp.(b). These results show that enabling the VLM to forecast future world evolution provides stronger safety awareness and planning efficiency, ultimately producing reliable autonomous driving behavior.


\begin{table}[t]
  \centering
  \small
  \caption{Ablation study on the world query of SGDrive.}
  \setlength{\tabcolsep}{1.pt}
  \vspace{-5pt}
  \begin{tabular}{
    c
    c c c c c c c
    c c c c c | c c
  }
    \toprule
    Exp. 
      & Scene
      & Agent
      & Goal
      & Furture
      & NC$\uparrow$  
      & DAC$\uparrow$ 
      & TTC$\uparrow$ 
      & EP$\uparrow$  
      & PDMS$\uparrow$  \\
    \midrule
    a 
      & \cmark & \xmark & \xmark & \xmark & 98.2 & 94.1 & 94.4 & 80.2 & \cellcolor{gray!30} 86.0  \\
    b 
      & \cmark & \cmark & \xmark & \xmark
      & 98.3 & 94.5 & 94.8 & 80.4 & \cellcolor{gray!30} 86.3  \\
    c 
      & \cmark & \cmark & \cmark & \xmark
       & 98.5 & 94.9 & 95.1 & 81.2& \cellcolor{gray!30} 87.0  \\
    d 
      & \cmark & \cmark & \cmark & \cmark
       & 98.6 & 95.1 & 95.4 & 81.2& \cellcolor{gray!30} 87.4  \\
    \bottomrule
  \end{tabular}
  \vspace{-5pt}
  \label{tab:ablation_dit}
\end{table}

\noindent\textbf{Ablation of world query for downstream planning.}

To assess the effectiveness of each subquery within our \textlangle\text{world}\textrangle queries, we conduct ablation studies on the downstream trajectory planning task using the stage~2 diffusion planner, and the result is shown in Table~\ref{tab:ablation_dit}. 
Exp.(a) employs only the scene-geometry layout to modulate trajectory generation, achieving a PDMS of 86.0. 
When safety-critical agents information is added, notable improvements are observed in key metrics such as NC and DAC, Exp.(b). 
Exp.(c) add driving goals into condition further leads to a significant enhancement in EP, indicating that activating high-level semantic intent effectively improves driving efficiency. 
Finally, incorporating future world-state predictions to guide the planner yields consistent gains across multiple metrics, resulting in a PDMS of 87.4. 
The additional improvements in TTC and NC further demonstrate that modeling the evolution of future scenes and road-user motions enables the planner to better anticipate potential hazards and avoid collisions.
These comprehensive results demonstrate that our proposed scene-agent-goal hierarchical cognition framework provides effective world knowledge guidance and substantially enhances overall driving performance.

\begin{figure*}
  \centering
  \vspace{-5pt}
  \includegraphics[width=1.0\textwidth]{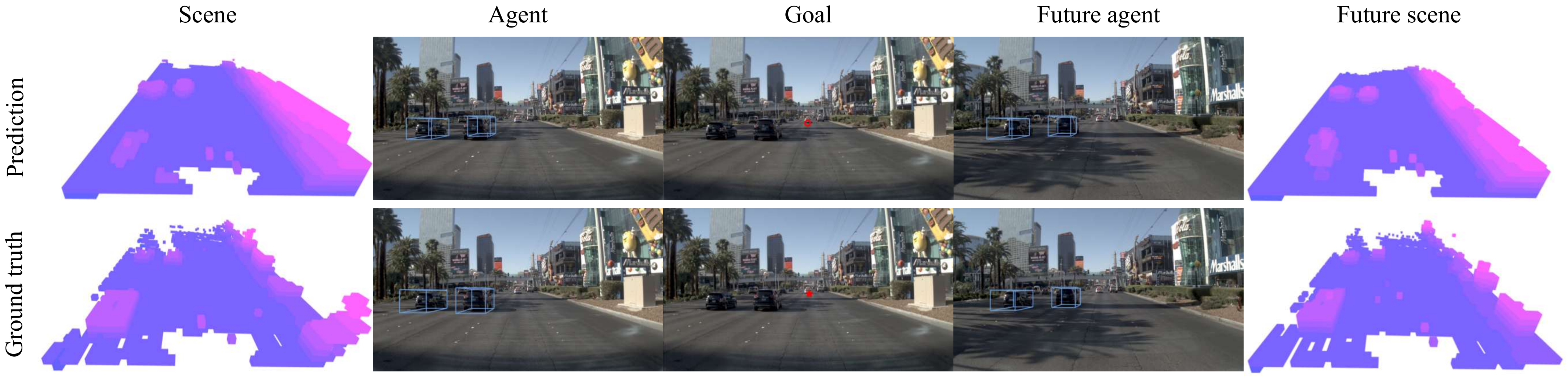} 
  \vspace{-5pt}
  \caption{Qualitative visualization of our model's predictions (top row) versus the ground truth (bottom row). The visualization shows our model accurately forecasts these hierarchical states, which closely align with the ground truth.}
  \label{fig:comparison_prediction_gt}
  \vspace{-5pt}
\end{figure*}

\begin{figure}
  \centering
  \includegraphics[width=1.0\linewidth]{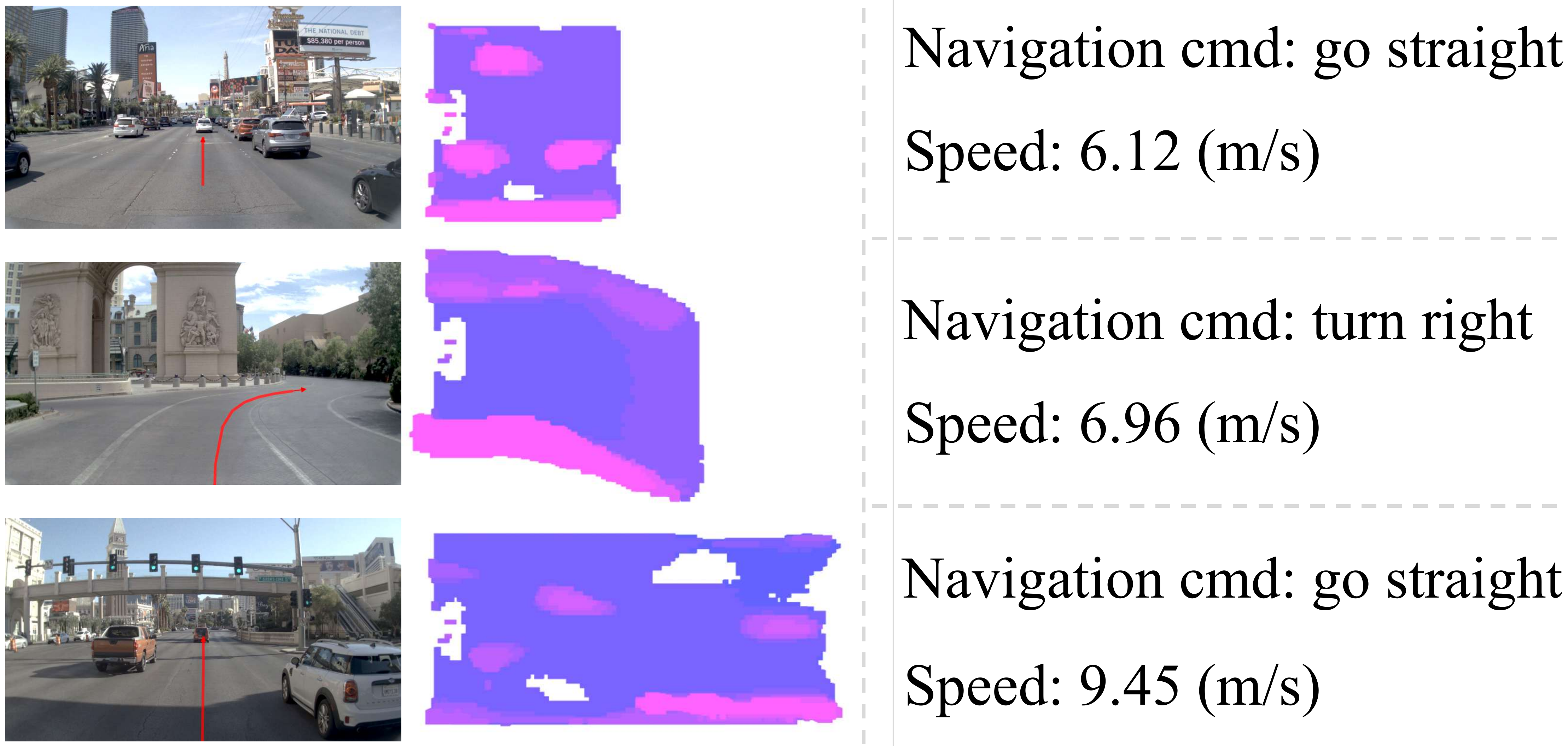}
  \vspace{-10pt}
  \caption{Ego-motion based adaptive geometric scene perception.}
  \label{fig:Adaptive_occ}
  \vspace{-5pt}
\end{figure}

\begin{table}[t!]
    \centering
    \small
    \caption{Ablation study on the attention mask of SGDrive.} 
    \vspace{-5pt}
    \begin{tabular}{@{}lccccccc@{}}
        \toprule
        Method & NC$\uparrow$ & TTC$\uparrow$ & EP$\uparrow$ & PDMS$\uparrow$ \\ 
        \midrule
        Causal  & 98.4 & 95.6  & 80.1 &\cellcolor{gray!30} 87.1 \\ 
        Structure  & 98.6 & 95.4  & 81.2 & \cellcolor{gray!30} 87.4 \\
        \bottomrule
    \end{tabular}
    \vspace{-5pt}
    \label{tab:ablation_results_atten_mask}
\end{table}

\noindent\textbf{Effect of structured attention masking.}

We compare our attention mechanism with default causal-attention method, as shown in Figure~\ref{fig:attention}. Causal-attention exposes each world query to all preceding tokens, introducing cross-category noise and semantic leakage; this corrupts world representations and causes the vehicle to adopt overly conservative behaviors~(e.g., slowing excessively to avoid potential collisions), thereby reducing driving efficiency. By contrast, our structured attention restricts visibility to same-type information, producing cleaner, task-specific embeddings. As shown in Table~\ref{tab:ablation_results_atten_mask}, this design yields a favorable trade-off: it improves EP, yielding a higher overall PDMS and more realistic driving behavior.

\subsection{Qualitative Results.}
\noindent\textbf{Compare with previous method.}
We select two representative scenarios to qualitatively compare our method with the previous state-of-the-art, RecogDrive~\cite{ReCogDrive}, as shown in Figure~\ref{fig:comparison_on_diff_model}. 
In the first scenario, involving multiple road users, RecogDrive’s predicted trajectory deviates significantly from the ground truth, leading to a potential collision. In contrast, our \netName{} empowered by explicit safety-critical agent detection, generates an optimal and collision-free trajectory. 
In the second, relatively open yet curved-road scenario, RecogDrive’s prediction drifts out of the lane and collides with the roadside barrier. Our model, however, accurately perceives the scene’s geometric layout and successfully avoids the collision. These results demonstrate that \netName{} effectively learns structured driving-world knowledge and can extrapolate it reasonably to ensure safe and rational driving behavior.

\noindent\textbf{Explicit world knowledge representation.}
In Figure~\ref{fig:comparison_prediction_gt}, we provide a qualitative comparison between our model’s predictions and the ground-truth annotations. The visualizations show a strong alignment across the scene-agent--goal hierarchy. This consistency indicates that our model has learned rich driving-world knowledge, enabling reliable perception and representation of both the current state and its short-horizon future evolution.

\noindent\textbf{Adaptive geometric scene perception.}
As shown in the Figure~\ref{fig:Adaptive_occ}, \netName{} adaptively perceives the driving scene according to the ego-vehicle's motion state and navigation command. For instance, at high speeds, it expands its perceptual horizon, whereas during turning maneuvers, it redirects its perceptual focus toward the turning direction. 
This demonstrates a more structured and effective representation of driving-relevant world knowledge, providing strong evidence that \netName{} successfully elicits the VLM's world-modeling ability.

\section{Conclusion}
We introduce \netName, a novel framework that explicitly structures a VLM's representation learning around a driving-specific knowledge hierarchy, enabling safer and more reliable autonomous driving. Our method leverages a scene-agent-goal hierarchical cognition scheme that disentangles the understanding of driving environments: it models the current world through scene geometry, road users, and driving goals, and further extrapolates their evolution into the future.
To support this, we design a structured attention-mask mechanism that prevents information leakage and suppresses cross-category noise. Finally, by integrating a DiT-based planner, our approach uses the inferred driving-world knowledge to regulate trajectory generation. Comprehensive experiments on NAVSIM demonstrate the effectiveness of our method and show that it achieves state-of-the-art performance in safe driving.

\newpage
{
    \small
    \bibliographystyle{ieeenat_fullname}
    \bibliography{main}
}

\clearpage
\setcounter{page}{1}
\maketitlesupplementary


\section{More experiments}
\label{sec:more_experiments}

\subsection{Comparison of hidden state fusion methods in diffusion planner}
\begin{table}[t!]
    \centering
    \small
    \caption{Comparison of hidden state fusion methods in diffusion planner.} 
    \vspace{-5pt}
    \begin{tabular}{@{}lccccccc@{}}
        \toprule
        Exp. & NC$\uparrow$ & TTC$\uparrow$ & EP$\uparrow$ & PDMS$\uparrow$ \\ 
        \midrule
        (a)  & 98.2 & 95.0  & 80.6 & 87.1 \\ 
        (b)  & 98.1 & 95.1  & 79.7 & 86.9 \\
        (c) & \textbf{98.6} & \textbf{95.4}  & \textbf{81.2} &  \textbf{87.4} \\
        \bottomrule
    \end{tabular}
    \vspace{-5pt}
    \label{tab:supp_diffusion}
\end{table}
As shown in Table~\ref{tab:supp_diffusion}, we compare several strategies for fusing hidden states within the diffusion planner. 
Exp.~(a) incrementally injects the hidden states of different subqueries across successive cross-attention layers. 
Exp.~(b) assigns distinct cross-attention layers to different subqueries. 
Exp.~(c), which corresponds to our proposed design, concatenates all subquery hidden states and enables interaction at every cross-attention layer. 
All fusion strategies achieve strong performance, confirming that our subqueries encode rich driving-world knowledge and can effectively guide the trajectory generation process.

\noindent\textbf{NAVSIM metric.}
NAVSIM~\cite{navsim} scores driving agents in two steps. First, subscores in range $[0,1]$ are computed after simulation. Second, these subscores are aggregated into the PDM Score (PDMS) $\in[0,1]$. 
We use the following aggregation of subscores based on the official definition:

\begin{equation}
\label{eq:pdms}
\begin{aligned}
\textrm{PDMS} = \underbrace{\Bigg( { \prod_{m \in \{\texttt{NC}, \texttt{DAC}\}}} \texttt{score}_m \Bigg)}_{\text{penalties}} \times \\
\underbrace{\Bigg( \frac{\sum_{w \in \{\texttt{EP}, \texttt{TTC}, \texttt{C}\}} \texttt{weight}_w \times  \texttt{score}_w}{\sum_{w \in \{\texttt{EP}, \texttt{TTC}, \texttt{C}\}} \texttt{weight}_w }  \Bigg)}_{\text{weighted average}}.
\end{aligned}
\end{equation}

Subscores are categorized by their importance as penalties or terms in a weighted average. A penalty punishes inadmissible behavior such as collisions with a factor $<1$. The weighted average aggregates subscores for other objectives such as progress and comfort. 

\noindent\textbf{NAVSIM metric with extended PDMS.} 
Hydra-MDP++~\cite{Hydra-mdp} extends the original PDMS metric by incorporating additional aspects of driving performance, including Traffic Lights Compliance (TL), Lane Keeping Ability (LK), and Extended Comfort (EC), providing a more comprehensive evaluation of a method’s effectiveness.
Formally, the Extended PDM Score (EPDMS) is computed as:
\begin{equation}
\label{eq:epdms}
\begin{aligned}
\mathrm{EPDMS} &= 
\underbrace{\prod_{m \in \{\mathrm{NC}, \mathrm{DAC}, \mathrm{DDC}, \mathrm{TL}\}} S^m}_{\text{penalty terms}} \\
&\quad \times 
\underbrace{
\frac{\sum_{w \in \{\mathrm{EP}, \mathrm{TTC}, \mathrm{C}, \mathrm{LK}, \mathrm{EC}\}} \mathrm{weight}_w \cdot S^w}
{\sum_{w \in \{\mathrm{EP}, \mathrm{TTC}, \mathrm{C}, \mathrm{LK}, \mathrm{EC}\}} \mathrm{weight}_w}
}_{\text{weighted average of positive indicators}}.
\end{aligned}
\end{equation}

Here, the first term accumulates multiplicative penalties for safety-critical violations, while the second term computes a weighted average over positive performance indicators, providing a balanced assessment of driving quality and comfort.

\begin{figure*}[t]
  \centering
  \vspace{-5pt}
  \includegraphics[width=1.0\textwidth]{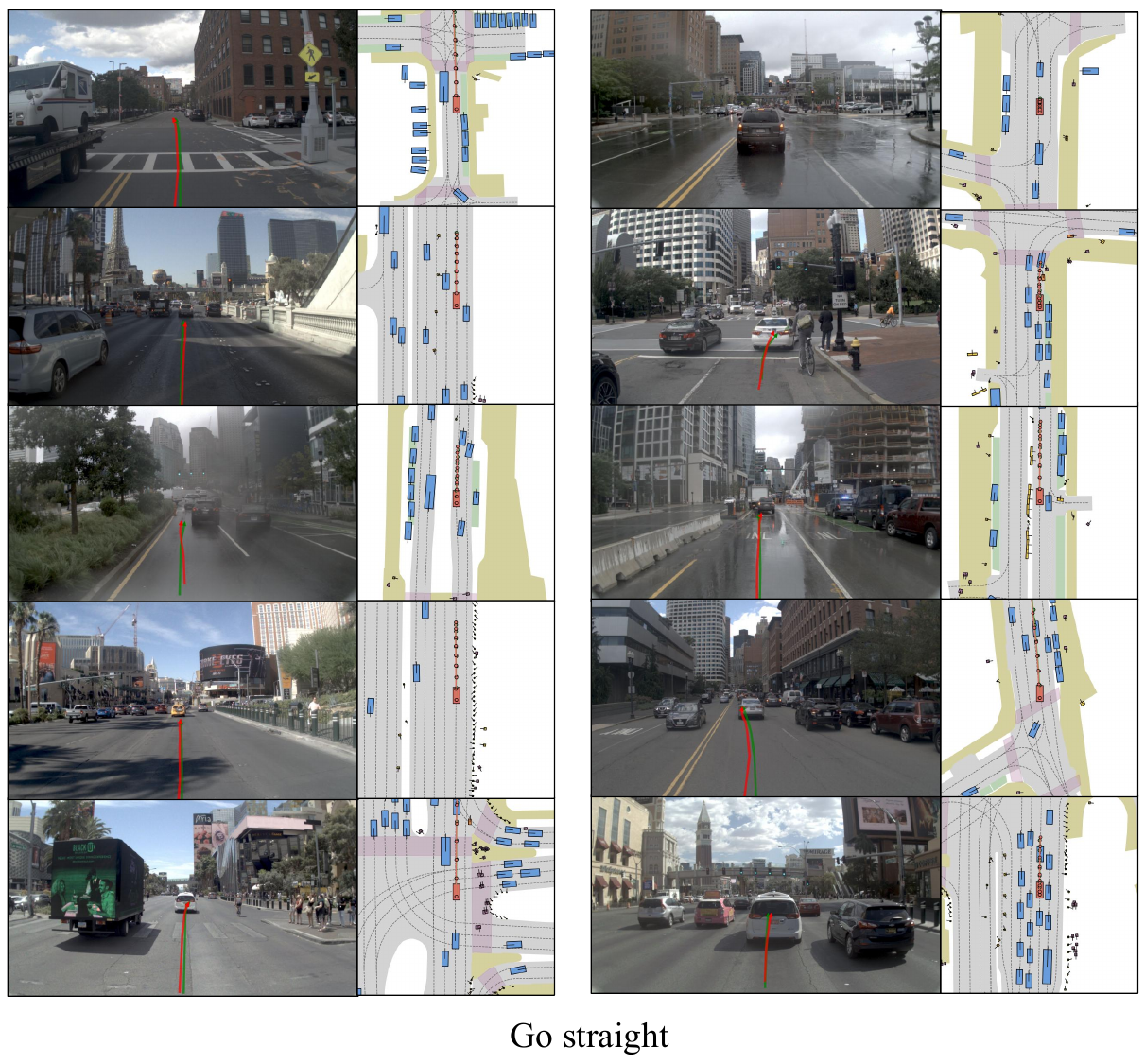} 
  \vspace{-15pt}
  \caption{Qualitative results on the Navtest benchmark.}
  \label{fig:supp_gostraight}
\end{figure*}
\section{Additional visualizations}
\subsection{Qualitative results}
\begin{figure*}[t]
  \centering
  \vspace{-5pt}
  \includegraphics[width=1.0\textwidth]{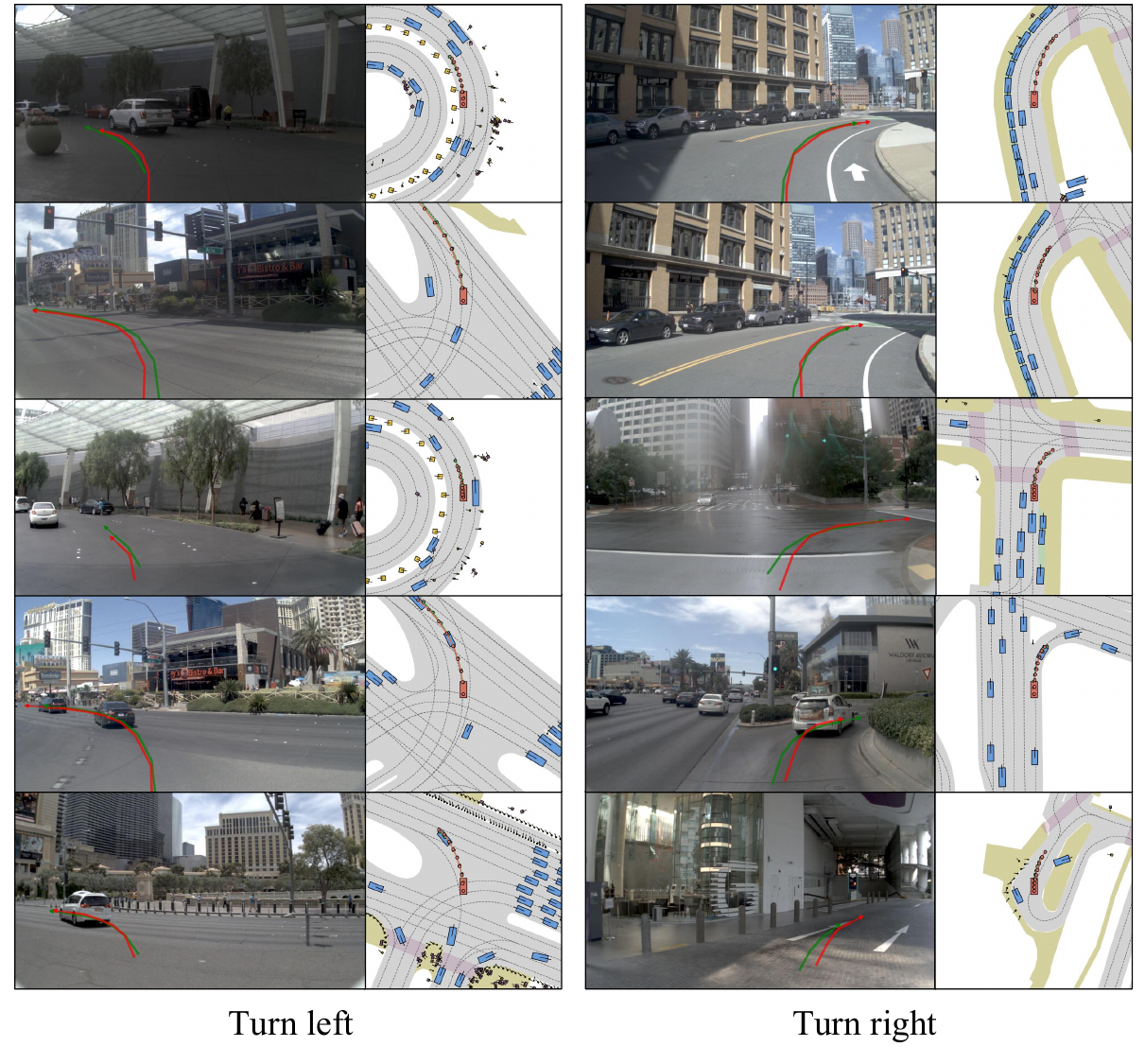} 
  \vspace{-15pt}
  \caption{Qualitative results on the Navtest benchmark.}
  \label{fig:supp_turn}
\end{figure*}

We provide additional qualitative results in Figure~\ref{fig:supp_gostraight} and Figure~\ref{fig:supp_turn}. For both straight-driving and turning scenarios, our predicted trajectories closely follow the ground truth. We also include several failure cases. 

\subsection{Failure cases}
\begin{figure*}[t]
  \centering
  \vspace{-5pt}
  \includegraphics[width=1.0\textwidth]{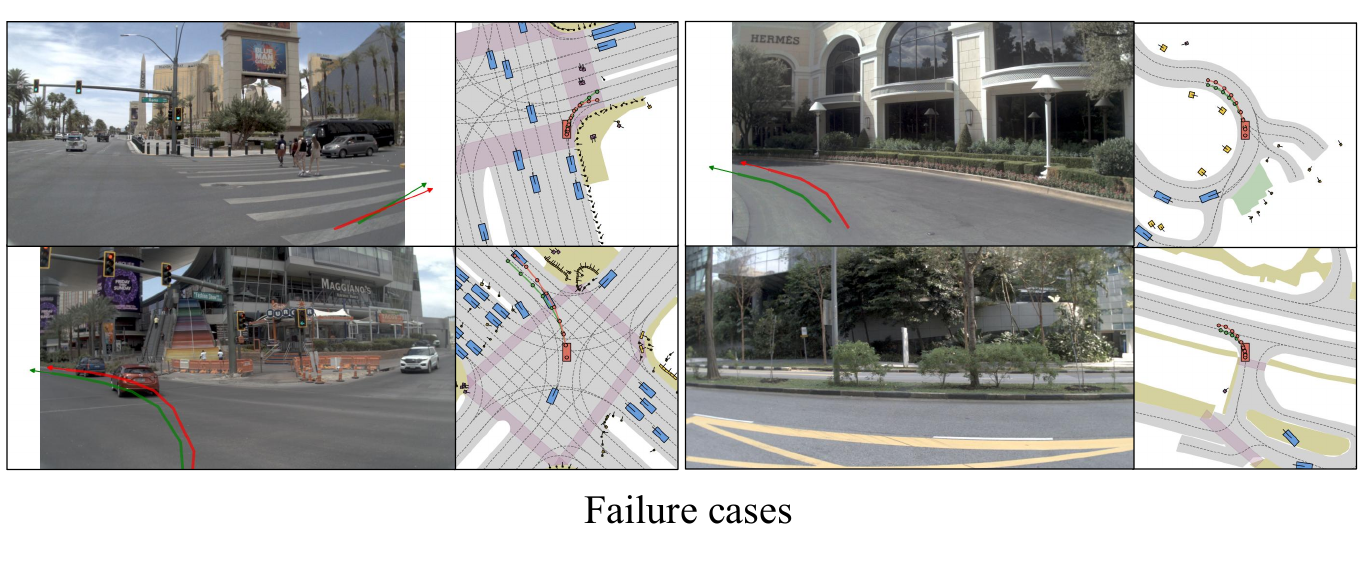} 
  \vspace{-15pt}
  \caption{Qualitative analysis of representative failure cases on the Navtest benchmark.}
  \label{fig:supp_fail}
\end{figure*}

As illustrated in Figure~\ref{fig:supp_fail}, when relying solely on a single front-view image, the model may exhibit slight deviations under extreme turning conditions. In such scenarios, due to the absence of corresponding viewpoints, accurately predicting long-horizon trajectories becomes challenging, sometimes leading to lane-change errors. Incorporating multi-view inputs is a promising direction to mitigate these limitations in future work.

\label{sec:rationale}

\end{document}